\documentclass[conference]{IEEEtran}

\usepackage{amssymb}
\usepackage{graphicx}

\usepackage{balance}  
\usepackage{graphics} 
\usepackage{times}    
\usepackage{url}      

\usepackage{array,multirow}
\usepackage{longtable}
\usepackage{tikz}
\usetikzlibrary{positioning}
\usepackage{booktabs}
\usepackage{todonotes}
\usepackage{algorithm}
\usepackage{algpseudocode}
\usepackage{amsmath}

\usepackage{algorithm, mathtools}
\usepackage{fancyhdr}

\newcommand{\code}[1]{{\lstinline!#1!}}

\newcommand{\argmax}{\operatorname*{arg\,max}}
\newcommand{\bff}[1]{{\bf{#1}}}

\def\S{{Section~}}

\usepackage{subcaption}
\makeatletter
\DeclareTextCommandDefault{\textleftarrow}{\mbox{$\m@th\leftarrow$}}
\makeatother

\makeatletter
\DeclareTextCommandDefault{\textrightarrow}{\mbox{$\m@th\rightarrow$}}
\makeatother

\begin{document}
%
\title{The N-Tuple Bandit Evolutionary Algorithm \\
for Automatic Game Improvement}


\author{\IEEEauthorblockN{Kamolwan Kunanusont, Raluca D. Gaina, Jialin Liu, Diego Perez-Liebana and Simon M. Lucas}
\IEEEauthorblockA{University of Essex, Colchester, UK\\
Email: \{kkunan, rdgain, jialin.liu, dperez, sml\}@essex.ac.uk}
}

\maketitle
 \thispagestyle{plain}
          \fancypagestyle{plain}{
            \fancyhf{} 
            \fancyfoot[L]{978-1-5090-4601-0/17/\$31.00~\copyright2017~IEEE} 
            \renewcommand{\headrulewidth}{0pt}
            \renewcommand{\footrulewidth}{0pt}
          }
\begin{abstract}
This paper describes a new evolutionary algorithm that is especially well suited to AI-Assisted Game Design. The approach adopted in this paper is to use observations of AI agents playing the game to estimate the game's quality. Some of best agents for this purpose are General Video Game AI agents, since they can be deployed directly on a new game without game-specific tuning; these agents tend to be based on stochastic algorithms which give robust but noisy results and tend to be expensive to run. This motivates the main contribution of the paper: the development of the novel N-Tuple Bandit Evolutionary Algorithm, where a model is used to estimate the fitness of unsampled points and a bandit approach is used to balance exploration and exploitation of the search space. Initial results on optimising a Space Battle game variant suggest that the algorithm offers far more robust results than the Random Mutation Hill Climber and a Biased Mutation variant, which are themselves known to offer competitive performance across a range of problems. Subjective observations are also given by human players on the nature of the evolved games, which indicate a preference towards games generated by the N-Tuple algorithm.
\end{abstract}

\IEEEpeerreviewmaketitle

\section{Introduction}

Automatic game design algorithms are systems capable of designing proper and playable games with close to none human intervention. Designing a ``playable'' game usually involves tuning an appropriate set of game parameters. Manually doing this might be time-consuming due to large search space of game parameters. 
Evolutionary Algorithms (EAs) are therefore employed to evolve game parameters, one of the first attempts being that of Togelius and Schmidhuber \cite{5035629}. Their results inspired the usage of EAs for game parameter tuning in later works.
This paper presents the results of a project that was solely focused on AI informed game design using three different Evolutionary Algorithms. The main aim is to explore the possibility of using AI controllers with different skill levels as human-player representations to evolve suitable game parameter sets using the same fitness criteria. In this context, a suitable game parameter set should be reflected in a game that better distinguishes player skill levels.  

To achieve this goal, a simple game \textit{Space Battle} was chosen and redesigned to \textit{Space Battle Evolved} by adding three new mechanics. A set of parameters of this game variant was selected (an interesting problem in itself) and evolved using a Random Mutation Hill Climbing algorithm \cite{BanditRMHC}, an improved version of RMHC called Biased-Mutation RMHC (B-RMHC) and a proposed noisy optimization N-Tuple Bandit Mutation. In order to reduce the dimensions of the game space, the rules were kept fixed and only several parameters of the game were tuned, having all $3$ Evolutionary Algorithms generate unique variants of \textit{Space Battle Evolved}. Several runs of automatic play testing were also carried out to ensure fine-tuned results, as well as human play testing to assess subjective game quality. 

This paper is organised as follows:
\S\ref{intro} briefly reviews the related work on automatic game design.
\S\ref{back} describes the game and AI controllers used in this paper.
\S\ref{appr} introduces our approach of evolving game instances using three Evolutionary Algorithms.
The experimental results are presented in \S\ref{xp} and \S\ref{conc} concludes the paper.

\section{Literature Review}\label{intro}

This section contains a brief review of several materials consulted as part of this research work.

Automatic game design is a sub-field of Game Artificial Intelligence that explores the idea of developing a system capable of generating dynamic and playable games. One of the first attempts at game design using such a system was developed by Togelius and Schmidhuber \cite{5035629}. The benefits of the work they pioneered include the possibility of creating multiple new and unique games automatically by making use of advanced computation methods and speed of execution. Their results suggested that evolutionary algorithms can indeed
be used to automatically search a space of possible games.

Nelson and Mateas \cite{Automated:AIIA07} used a generative process in their paper, which refers to factoring a game design process into four interacting domains: abstract game mechanics, game representation, thematic content and control mapping. The game design space, which is the space of all possible games that the resultant system can reason about, is defined by the specific knowledge given for each of these domains. 


The basic methodology of creating a generative system was employed by Isaksen et al. \cite{IsaksenGN15,IsaksenICCC15,AIIDE1511594}. They explored the possibilities of discovering useful variants of games by tuning aspects of the game space and analysing the resulting player experience. Their paper is focused on the possibilities of achieving this effect by varying game parameters without changing the game rules and how this process could yield games of varying difficulties. 

Another application is the Physical Traveling Salesman Problem map evolution. Perez et al. \cite{6605565} use three AI players of different skill levels to evaluate the maps produced by their algorithm, the hypothesis described in the paper being based on the fact that the players’ rankings would be kept consistent in a good map; the higher the skill depth, the better the map. A similar approach was used in the experiments carried out for this present paper, with a fitness calculation aiming to distinguish the between the skill levels of three AI players.

Additionally, research has looked at automated maze generation for Ms Pac-Man. Safak et al. \cite{SafakPM16} use genetic algorithms to this end, using as a measure of fitness the ability of a player to finish the game in the newly created map. Their results show that Evolutionary Algorithms can be used to generate interesting mazes quite different from the original design, offering the players new challenging experiences.

A detailed account of the all game design aspects through search-based Procedural Content Generation is given by Togelius et al. in \cite{Togelius2011PCG}. One thing highlighted in this survey is the importance of efficiently encoding the search space, not only for the EA to be able to process it correctly, but also for the evaluation function to analyse it effectively. To this extent, an emphasis is put on the necessity of search space constraints and a clear encoding in \cite{RothalaufEA}, which in turn would result in the same genotype being easily adapted to different phenotypes in other applications by simply varying the mapping functions. Therefore the work in the present paper focuses on describing the game space in terms of a set of interesting parameters with limited value ranges, adapted for fitness evaluations by inserting the generated values into games to be analysed.

Nielsen et al.~\cite{7317941} expand the automated design process to general video games and game rules, using the Video Game Description Language. A similar fitness measure as the one employed in our study is used, therefore comparing bad players against good players and using the difference in their performance to quantify the quality of the game. The authors analyse the differences in human-designed games, mutated variants of these games and newly generated games, reporting mixed results, with interesting but hardly playable games resulting from their study. They finally recommend evolutionary algorithms to be used only for idea generation and improvement, instead of actually creating complete games.

Menezes et al.~\cite{4100132} present their initial conceptual model incorporating a novel approach to generating engaging and surprising game worlds, based on the theory that complex things may emerge from simple interactions. They focus on co-evolution as the core of a Complex Adaptive System and encourage less human involvement in the evolutionary process, instead making use of auto-organizative systems. 

\section{Background}\label{back}

This sections offers a brief description of the games and AI controllers used in the experiments.

\subsection{Space Battle game}\label{game}

\begin{figure}[!t]
    \centering
        \includegraphics[width=0.49\columnwidth]{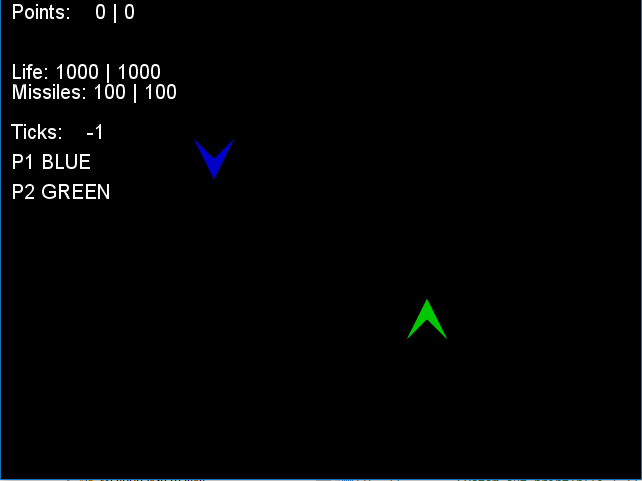}
    \includegraphics[width=0.49\columnwidth]{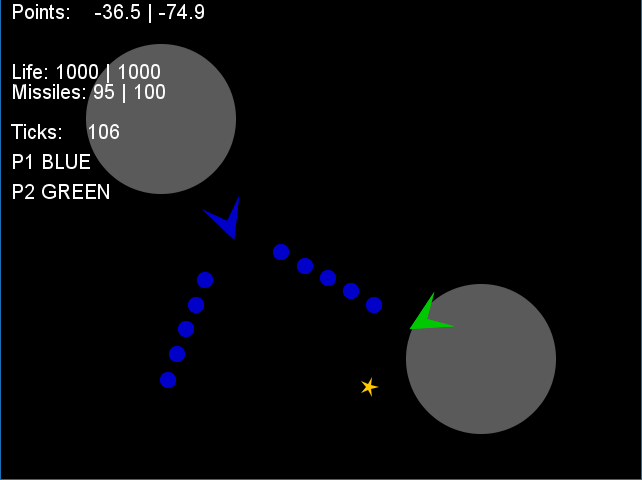}
          \caption{Space Battle (left) and Space Battle Evolved (right)}
          \label{fig:new_spacebattle} 
\end{figure}

The original Space Battle game is a 2-player competitive game wherein players pilot ships and aim to shoot their opponent. The ships are in convex-quadrilateral shapes, as shown in Figure \ref{fig:new_spacebattle}, with the front being indicated by a single acute angle point. The first player controls the blue ship while the second player’s ship is green. The starting positions of the player ships are as depicted in Figure \ref{fig:new_spacebattle} (left).

Both players act simultaneously and have the same range of actions available, these being: turn clockwise, turn anticlockwise, thrust (move forward) and shoot a missile. Rotation, acceleration and shooting actions can be performed together in one game tick. Turn actions simply rotate the ship into the corresponding directions, without moving it from its current position. Therefore, the ship can only move forward when the thrust action is operated. When a player chooses to shoot, a round-shaped missile appears at the player’s ship location and moves into the ship's forward direction with a specific velocity. 

Each player has $1000$ lives, which decrease by $1$ if their ship is hit by one of the opponent’s missiles. They each start with only $100$ missiles. The game ends when one of two constraints is met: one player’s lives dropping to zero or the game-end being reached after a set amount of game ticks ($2000$ by default). At the end of the game, the player with most points is declared the winner.

The framework uses the same interface as the Two-Player General Video Game AI (GVGAI) framework \cite{GVGAI,GVGAI2P}, making it easy to plug in agents submitted to the GVGAI competition and use them for game testing in this new problem. All AI controllers have access to a Forward Model (FM), which allows the agents to simulate possible future game states by providing an in-game action. Additionally, as the game is real-time, the agents only have $40$ms for making a decision during a game step, with $1$s for initialization.

This game appears in recent literature as an interesting challenge for AI agents. Liu et al. \cite{Liu2016} analyse a novel co-evolution approach to two player games and test it on a simplified Space Battle game, in which the shooting action is removed and instead players score by positioning behind the enemy ship. The game states are evaluated by calculating the distance the players are from the ideal position.

\subsection{AI controllers}\label{controller}

Six different game playing agents were used in this experiment for game testing purposes. These are separated into two sets: first, the enemy player set, containing all agents mentioned in this section. The enemy player is an evolvable parameter (see Section \ref{sec:params}), therefore the EA has access to all possible enemies when evolving the game.

In order to analyse the skill-depth of games produced by the Evolutionary Algorithm, a second set is used, containing only three players: One Step Look Ahead (1SLA), which delegates a non-skill player, Rotate and Shoot (RAS), portraying an intermediate-skill player and Monte Carlo Tree Search (MCTS), representing a skillful player. 

\subsubsection{Do Nothing}

Do Nothing is a dummy controller that returns no action in every game tick.

\subsubsection{Random}

The random controller returns a randomly chosen action in every time step, out of all those available to it. This agent can prove to be challenging to beat due to its unpredictable nature.

\subsubsection{One Step Look Ahead (1SLA)}

This is the most simple AI algorithm which accesses the Forward Model. One Step Look Ahead (1SLA) simply simulates the next state for all available actions and chooses for execution the most promising action. The agent uses a simple heuristic to quantify the value of a game state, by considering the game score and, in cases of end-game states, whether it won or lost (receiving either a large bonus or a large penalty, respectively). The general behaviour of this agent observed in the game used for this experiment is moving randomly without shooting, unless the missile is due to hit the other player in the next game tick. It was considered a low skilled player for this study because it is fairly easy to beat, as long as the other player keeps its distance and avoids being right in the missile spawn point.

\subsubsection{Rotate and Shoot (RAS)}

As the name suggests, this controller returns a combination of actions at each time step, clockwise turning and shooting. This strategy has proven to be unbeatable in a game which allows unlimited or a large number of missiles. In the game used in this paper, the players only have $100$ missiles available in the beginning. Therefore, in theory, RAS should not perform well after the first $100$ time steps. It was chosen as a mid-skilled AI player in this experiment, as it has a potential to survive and shoot the enemy at least in at the beginning of the game.

\subsubsection{Monte Carlo Tree Search (MCTS)}

Monte Carlo Tree Search (MCTS) is a well-known tree search algorithm for game AI controllers. The main strength of MCTS is its ability to deal with huge search space by balancing between known and non-explored states using UCB1 equation (See Equation~\ref{eq:ucb1}). For more information on this technique, the reader is referred to \cite{MCTSsurvey}. MCTS was chosen to represent a skillful player because its behaviour in the game was mostly unpredictable, unlike 1SLA and RAS, and able to perform well for the whole duration of the game.

\begin{equation}\label{eq:ucb1}
a^* = \argmax_{a \in A(s)} \left\{Q(s,a) + C \sqrt{\frac{ \ln N(s) }{ N(s,a) }}\right\}
\end{equation}

\subsubsection{Microbial Evolutionary Algorithm (MEA)}

This controller picks the first action of the best action plan generated by a Evolutionary Algorithm. It starts with a population of random individuals encoded as sequences of in-game actions. It then selects individuals through a microbial tournament, from which an offspring is generated via crossover. The newly generated individual is mutated, its fitness evaluated (using the same heuristic as 1SLA on the game state reached after playing the actions in one individual, in order) and the best individuals are carried forward to the next generation. This process is reiterated until the budget limit is reached (e.g. time, memory or specific number of iterations).

\subsection{Random Mutation Hill Climber (RMHC)}

A Random Mutation Hill Climber is the simplest version of an evolutionary algorithm, with only one individual in the population. It starts by randomly assigning values to each gene of the individual. One gene is selected for mutation uniformly at random, the fitness value of the resulting individual is calculated (similarly to the MEA evaluation) and compared with the previous one. The better individual is kept for the next iteration of the algorithm, repeated while allowed by the budget offered. In the implementation used for this paper, both the parent and the offspring are evaluated in each generation.

This algorithm is often used as a game playing agent in literature due to the great results produced while keeping simplicity. Buzdalov et al. \cite{6784612} use an RMHC algorithm combined with Q-Learning for adaptive behaviour and analyse its runtime complexity on a modified OneMax problem (with an obstructive fitness function meant to lead the algorithm in the wrong direction), reporting good results. Liu et al. \cite{BanditRMHC} explore modifications of the vanilla RMHC method, by using Upper Confidence Bounds (UCB) to guide evolution.  

\section{Approach}\label{appr}

Three EA algorithms were used to evolve \textit{Space Battle Evolved} game parameters, using a fitness function meant to distinguish between good and bad players and optimize games in order to maximise the skill depth (See Section \ref{section:fitness_eval}). 

For each algorithm, $50$ trials were experimented, due to the noisy game environment, as the same parameter set might return different fitness values in different runs. Each iteration of the evaluation process was carried out for $100$ evaluations. The final games were analysed statistically by their fitness, as well as tested by a number of human players who offered their subjective opinions.

\subsection{Space Battle Evolved}\label{newgame}

\textit{Space Battle Evolved} (See Figure \ref{fig:new_spacebattle} for an example) is a variation of the simple Space Battle game that was designed for this project. Following on from the original rules of Space Battle, there are three main changes made to produce this variant.

Firstly, black holes were created, which have a set range and add forces to nearby objects in order to drag them towards their center. The players receive a penalty for each game tick in which they remain within a certain distance to a black hole center. There are, however, areas inside the black holes where no penalty is applied, called safe zones.

Secondly, two additional types of missiles were included in this version: a twin shot type, which fires two normal missiles at $45$ and $-45$ degree angles from its direction and a bomb type which explodes in a large radius after a set time or upon collision with another object.

Lastly, due to the limited missiles available, the players have now at their disposal collectible packs of $20$ missiles, which spawn on the map in a random position, disappear after a specified time (or after being collected by a player) and re-spawn again after a certain amount of time.

\subsection{Evolvable game parameters}\label{sec:params}

\begin{table*}[!t]
\centering
\caption{Evolvable parameters, their value ranges and step.}
\label{tab:params}
\resizebox{\textwidth}{!}{%
\begin{tabular}{|c|c|c||c|c|c||c|c|c|}
\hline
\textbf{Parameter}  & \textbf{Value Range} & \textbf{Step} & \textbf{Parameter}   & \textbf{Value Range} & \textbf{Step} & \textbf{Parameter} & \textbf{Value Range} & \textbf{Step} \\ \hline\hline
MISSILE\_MAX\_SPEED & 1 - 10               & 1             & BLACKHOLE\_CELL x 16 & 0 or 1               & 1             & BOMB\_RADIUS       & 10 - 50              & 10            \\ \hline
MISSILE\_COOLDOWN   & 1 - 9                & 1             & BLACKHOLE\_RADIUS    & 25 - 200             & 25            & MISSILE\_TYPE      & 0 - 2                & 1             \\ \hline
MISSILE\_RADIUS     & 2 - 10               & 2             & BLACKHOLE\_FORCE     & 0 - 3                & 1             & RESOURCE\_TTL      & 400 - 600            & 100           \\ \hline
MISSILE\_MAX\_TTL   & 40-160               & 20            & BLACKHOLE\_PENALTY   & 0 - 9                & 1             & RESOURCE\_COOLDOWN & 200 - 300            & 50            \\ \hline
GRID\_SIZE          & 1 - 4                & 1             & SAFE\_ZONE           & 0 - 20               & 10            & ENEMY\_ID          & 0 - 5                & 1             \\ \hline
\end{tabular}%
}
\end{table*}

There were $30$ evolvable game parameters in total, as summarized in Table \ref{tab:params}. These can be divided into $4$ categories: missiles, black holes, resources and enemy.

\subsubsection{Missile related}

$6$ of the parameters refer to missiles, including the type of the missile, its maximum speed, its cooldown (how many game ticks until the player is allowed to shoot a new missile), its radius, its time to live and, finally, the bomb explosion radius (for bomb type missile only). As the primary way of obtaining a score advantage over the opponent (and thus possibly ultimately winning the game), these could be considered key parameters.

\subsubsection{Black hole related}

$21$ of the game parameters involve black holes, $17$ relating to black hole locations and the rest specifying black hole characteristics. We divided the game map into a grid and allowed the evolutionary algorithms to decide whether to include a black hole in the center of each grid cell or not. The grid size varies between 1 and 4 (with a step of 1), therefore there could be up to $16$ ($4 \times 4$) black holes in a game. The black holes layout depict the main environment of the game. 
Other parameters needed for black hole mechanics include radius, force, penalty score (negative score given when the player is inside the black hole) and the safe-zone radius (non-penalization area around the inside border of the black hole).


\subsubsection{Resource related}

This game object refers to a pack of $20$ missiles which eventually spawn on random positions on the map for collection. $2$ resource-related parameters are time to live (number of game ticks before the resource disappears) and cooldown (number of game ticks for the resource to be re-spawned). Due to the limited missiles in \textit{Space Battle Evolved}, the resources dictate whether the players may gain more than the maximum possible $10000$ points which can be achieved with the initial missile budget.

\subsubsection{Enemy related}

The algorithms presented in this paper evolve the AI enemy as part of the game parameters, a novel aspect which greatly impacts the resulting gameplay. This parameter can refer to either of the AI controllers mentioned in Section \ref{controller}.

\subsection{Baseline algorithm}

We applied Random Mutation Hill Climber as a baseline algorithm. The algorithm uses an array of $30$ parameters for evolution, initializing each one to a random value. One parameter is then chosen uniformly at random and mutated ($1$ random gene being changed to a different value). The fitness value of this mutated game is calculated by playing three games with the three AI controllers of different skill levels and following the method in Section \ref{section:fitness_eval}. If the mutated game ends up with a higher fitness than its parent, the offspring is the individual carried forward to the next generation. RMHC algorithm implementation details can be observed in Algorithm \ref{alg:RMHC}.

\begin{algorithm}[!t]
\caption{Random Mutation Hill Climber (RMHC)}\label{alg:RMHC}
\begin{algorithmic}[1]
\small
\State $\text{{Input: game parameter list \textit{params}, number of trials \textit{ntrials},}}$
\State\hspace{0.75cm} $\text{{number of evaluations \textit{nEvals}}}$
\State $\text{{Output: evolved parameter sets}}$
\State $\text{{BEGIN}}$
\State\hspace{0.5cm} $\text{\textit{paramList} \textleftarrow  \hspace{0.075cm}$\varnothing$}$
\State\hspace{0.5cm} $\text{REPEAT \textit{ntrials} times}$
\State\hspace{1.0cm} $\text{\textit{params}  \textleftarrow  \hspace{0.075cm} randomly assign each value}$
\State\hspace{1.0cm} $\text{\textit{bestSoFar}  \textleftarrow  \hspace{0.075cm} \textit{fitness}(\textit{params})}$
\State\hspace{1.0cm} $\text{REPEAT \textit{nEvals} times}$
\State\hspace{1.5cm} $\text{{\textit{p} \textleftarrow \hspace{0.075cm} randomly select one parameter}}$
\State\hspace{1.5cm} $\text{{\textit{mutatedParams} \textleftarrow \hspace{0.075cm} \textit{mutate}(\textit{params,p})}}$
\State\hspace{1.5cm} $\text{\textit{newFitness}  \textleftarrow  \hspace{0.075cm} \textit{fitness}(\textit{mutatedParams})}$
\State\hspace{1.5cm} $\text{IF \textit{newFitness} $\geq$ \textit{bestSoFar}}$
\State\hspace{2.0cm} $\text{\textit{params}  \textleftarrow  \hspace{0.075cm} \textit{mutatedParams}}$
\State\hspace{2.0cm} $\text{\textit{bestSoFar}  \textleftarrow  \hspace{0.075cm} \textit{newFitness}}$
\State\hspace{1.0cm} $\text{add \textit{params} to \textit{paramList}}$
\State\hspace{0.5cm} $\text{{RETURN \textit{paramList}}}$
\State $\text{{END}}$
\end{algorithmic}
\end{algorithm}

\subsection{Biased Mutation RMHC}

Biased Mutation RMHC (B-RMHC) was inspired by the idea that different parameters affect the change in fitness values at different rates. That is, modifying one parameter might significantly affect the fitness value more than others. Therefore, a biased mutation towards more interesting parameters was used to obtain more diverse games and speed up evolution. The algorithm is made up of two parts: parameter pre-processing and actual evolution.

\subsubsection{Pre-processing}

The parameters were divided into two groups, separating the black hole cells (Group B) from the rest (Group A). For Group A's pre-processing, the parameters received random values to start with. Then, for each parameter, the \textit{importance} metric was calculated by using the standard deviation from the fitness in $N$ tests, where $N$ is the total number of values the parameter tested can take. For each value, the game taking the new parameter list was evaluated using the same fitness function employed during evolution. This assessment is based on the assumption that larger differences in fitness lead to larger standard deviation values. 

For Group B, the parameters were analyzed separately for each possible grid size value, starting from all the black hole cells being empty and evaluating the effect of enabling a black hole in each cell. Similarly to Group A, the standard deviation of the fitness values resulted from each cell's evaluation was used to rank these parameters.

Pre-processing step outputs two list of parameters sorted by how much they affect the game fitness value. Details of this algorithm can be seen in Algorithm \ref{alg:mut_prep}. This ordering was then used in the evolution step.

\begin{algorithm}[!t]
\caption{Biased Mutation Pre-processing (MutPrep)}\label{alg:mut_prep}
\begin{algorithmic}[1]
\small
\State $\text{{Input: game parameter list \textit{params}}}$
\State $\text{{Output: sorted lists of important parameters}}$
\State$\text{{BEGIN}}$
\State\hspace{0.5cm} $\text{{\textit{PriorityQParams} \textleftarrow \hspace{0.075cm} $\varnothing$}}$
\State\hspace{0.5cm} $\text{{\textit{PriorityQBH} \textleftarrow \hspace{0.075cm} $\varnothing$}}$
\State\hspace{0.5cm} $\text{{\textit{ParamsN} \textleftarrow \hspace{0.075cm} \textit{GroupA} parameters}}$
\State\hspace{0.5cm} $\text{{FOR EACH \textit{p} in \textit{paramsN}}}$
\State\hspace{1.0cm} $\text{\textit{value} \textleftarrow \hspace{0.075cm}$\varnothing$}$
\State\hspace{1.0cm} $\text{\textit{rand}\textleftarrow \hspace{0.075cm}randomly assign other parameter values}$
\State\hspace{1.0cm} $\text{FOR EACH possible value \textit{v} of \textit{p}}$
\State\hspace{1.5cm} $\text{\textit{rand}[\textit{p}] \textleftarrow \hspace{0.075cm} \textit{v}}$
\State\hspace{1.5cm} $\text{add \textit{fitness}(\textit{rand}) to \textit{value}}$
\State\hspace{1.0cm} $\text{\textit{PriorityQParams}[\textit{p}] \textleftarrow \hspace{0.075cm} \textit{SD}(\textit{value})}$
\State\hspace{0.5cm} $\text{\textit{rand}\textleftarrow \hspace{0.075cm}randomly assign other parameter values}$
\State\hspace{0.5cm} $\text{\textit{rand}\textleftarrow \hspace{0.075cm}disable all black holes}$
\State\hspace{0.5cm} $\text{FOR \textit{gSize} $\in$ $\{0,1,2,3,4\}$}$
\State\hspace{1.0cm} $\text{\textit{bhpriorityQ} \textleftarrow \hspace{0.075cm}$\varnothing$}$
\State\hspace{1.0cm} $\text{\textit{fitnessOff} \textleftarrow \textit{fitness}(\textit{rand})}$
\State\hspace{1.0cm} $\text{FOR \textit{b} = 1 to \textit{$gSize^2$}}$
\State\hspace{1.5cm} $\text{enable black hole at position \textit{b} in \textit{rand}}$
\State\hspace{1.5cm} $\text{\textit{bhpriorityQ}[\textit{b}] \textleftarrow \hspace{0.075cm}\textit{fitnessOff} $-$ \textit{fitness}(\textit{rand})}$
\State\hspace{1.0cm} $\text{\textit{PriorityQBH}[\textit{gSize}] \textleftarrow \hspace{0.075cm}\textit{bhpriorityQ}}$
\State\hspace{0.5cm} $\text{{RETURN \textit{PriorityQParams}, \textit{PriorityQBH}}}$
\State $\text{{END}}$
\end{algorithmic}
\end{algorithm}

\subsubsection{Evolution}

A softmax function was employed to do a biased parameter selection at the beginning of the evolution process. This ensures that more important parameters are more likely to be selected. After that, the algorithm follows the design of simple RMHC and the fitness is evolved for a fixed number of iterations.



\subsection{N-Tuple Bandit Evolutionary Algorithm}

The N-Tuple Bandit Evolutionary Algorithm is an algorithm we developed to be particularly useful for evolving game designs and game parameters, especially when using agent-based evaluation methods. 
The evaluation function used in this paper will be noisy if the game is played by stochastic agents, such as MCTS, and fairly expensive in CPU time to run each game. Hence, it is desirable to have an evolutionary algorithm that is able to operate very efficiently, making the best possible use of the available fitness evaluation budget, and also one that is robust to noise.
The N-Tuple Bandit EA satisfies these criteria.  

\subsubsection{Algorithm}

The algorithm operates as follows. It begins by choosing a random point in the search space, which is called the current point. 
It then makes a noisy fitness evaluation and stores it in the N-Tuple Fitness Landscape Model as the value for that point. Using a mutation operator to generate a set of unique neighbours of the current point, and using the fitness landscape model, the algorithm gets the estimated Upper Confidence Bound (UCB) value of each point (see Equation \ref{eq:ucb1}). Finally, it sets the current point as the neighbour from the previous step with the highest UCB value. These steps are displayed in Algorithm \ref{alg:ntuple}. When the fitness evaluation budget has been exhausted, the method searches a set of neighbours of all of the evaluated points and returns the one with the highest mean value ($Q(s,a)$). 
\begin{algorithm}[!t]
\caption{N-Tuple Bandit Mutation (NTuple)}\label{alg:ntuple}
\begin{algorithmic}[1]
\small
\State $\text{{Input: game parameter list \textit{params}}}, $
\State\hspace{0.75cm} $\text{{number of evaluations \textit{nEvals}}}$
\State $\text{{Output: the best parameter set}}$
\State $\text{{BEGIN}}$
\State\hspace{0.5cm}
$\text{\textit{current} \textleftarrow \hspace{0.075cm} randomly assign each value}$
\State\hspace{0.5cm}
$\text{\textit{LModel}  \textleftarrow \hspace{0.075cm} $\varnothing$}$
\State\hspace{0.5cm}
$\text{REPEAT \textit{nEvals} times}$
\State\hspace{1.0cm}
$\text{\textit{value} \textleftarrow \hspace{0.075cm} \textit{fitness}(\textit{current})}$
\State\hspace{1.0cm}
$\text{add $<$\textit{current}, \textit{value}$>$ to  \textit{LModel}}$
\State\hspace{1.0cm}
$\text{\textit{neighbors} \textleftarrow \hspace{0.075cm} generate neighbours from \textit{LModel}}$
\State\hspace{1.0cm}
$\text{\textit{current} \textleftarrow \hspace{0.075cm} \textit{n} in \textit{neighbors} with \textit{Max(\textit{UCB}(\textit{n}))}}$
\State\hspace{0.5cm}RETURN \textit{n} in \textit{LModel} with highest average value
\State $\text{{END}}$
\end{algorithmic}
\end{algorithm}

\subsubsection{N-Tuple Fitness Landscape Model}

N-Tuple systems have ideal properties for use as fitness landscape models, in that they offer super-fast one-shot training and good accuracy.
While their use for optical character recognition dates back to the 1950s, Lucas \cite{NTupleOthello} introduced their use for game position evaluation functions.
The concept is as follows. Given an $D$-dimensional search space, we sub-sample its dimensions with a number of $N$-tuples. The value of $N$ ranges from $1$ up to $D$, though may miss out values in between. The results in this paper are based on using $D$ 1-tuples and 1 $D$-tuple.

Each N-Tuple has a look-up table (LUT) that stores statistical summaries of the values it encounters; the basic numbers stored are the number of samples, the sum of the samples, and the sum of the square of the samples. This enables the mean, the standard deviation and the standard error to be calculated for each entry in the table. 
More details can be found in \cite{NTupleOthello}.

\subsection{Fitness evaluation}\label{section:fitness_eval}

The fitness value of each game was evaluated with 3 gameplays, by using 1SLA, RAS and MCTS as players. For each game played, both of the players scores were divided by $100$ to lower the scale, then a $1000$ bonus points were awarded to the winner to prioritize winning result in producing the final score. The difference in the final score between player 1 and player 2 was assigned as the fitness of one game. Equation \ref{eq:total_Score} shows the final score calculation for each gameplay. $W_k$ = $1000$ if the player $k$ won the game and $0$ otherwise.

 \begin{equation}
 \label{eq:total_Score}
T_{g} = {(\frac{S_1}{100}+{W_{1}})} - {(\frac{S_2}{100}+{W_{2}})}
 \end{equation}
 
After the total score $T_g$ for every game $g$ is computed, it was brought into the final fitness calculation as depicted in Equation \ref{eq:fitness}, where $T_1$ is the weak player's game fitness (1SLA), $T_2$ is the mediocre player's game fitness (RAS) and $T_3$ is the strong player's game fitness (MCTS).

 \begin{equation}
  \label{eq:fitness}
Fitness = Min(T_{3}-T_{2}, T_{2}-T_{1})
 \end{equation}
 
Equation \ref{eq:fitness} is similar to that used for the Physical Traveling Salesman Problem by Perez et. al. \cite{6605565}. Based on this fitness evaluation, the aim of the algorithms is to maximize the smallest gap between final scores of each game in the order $T_{3}$  \textgreater $T_{2}$  \textgreater   $T_{1}$, which would result in the maximum skill-depth.

 
\section{Experimental results}\label{xp}
We apply the RMHC, the Biased Mutation RMHC (denoted as B-RMHC) and the N-Tuple Bandit Mutation algorithm (denoted as N-Tuple) independently $50$ times to evolve game instances, thus $150$ games are designed in total. 
100 game evaluations are allocated to each of the algorithms during the evolution. Then we pick up some of the evolved game instances for human players to test and analyse their feedback.

\subsection{Selection of designed games by reevaluation}
\begin{figure*}[ptb]
\centering
\includegraphics[width=0.8\textwidth]{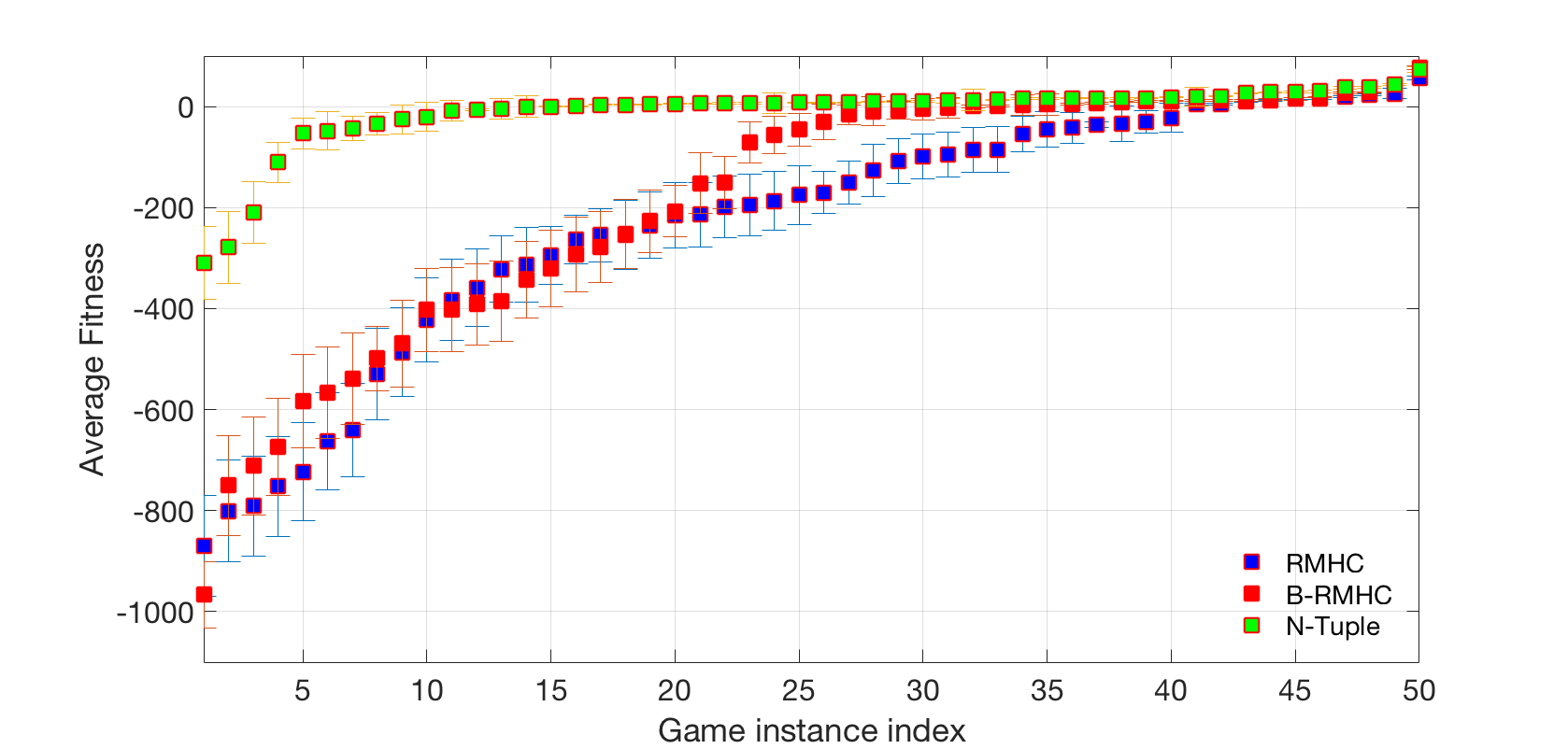}
\caption[]{\label{valid}Sorted average fitness values over 100 evaluations of 50 game instances evolved using three different algorithms. The x-axis shows the game indices after sorting. The standard errors are shown as well.}
\end{figure*}

To select the game instance for human testing, each of the evolved game instances is then evaluated $100$ times, where each evaluation takes into account the outcomes of three games played by the 1SLA, RAS and MCTS controllers (detailed in \S\ref{section:fitness_eval}). The sorted average fitness values over $100$ evaluations and standard errors are presented in Figure \ref{valid}.

The N-Tuple algorithm (green markers) outperforms both the RMHC and its variant. Moreover, N-Tuple is more robust and has a more stable performance (negligible standard error). Among $50$ game instances evolved by the N-Tuple Bandit Mutation algorithm, only a few of them (very left part in Figure \ref{valid}) have an average fitness below zero. Nevertheless, the lowest average fitness is still much higher than most of the games evolved by both RMHC and B-RMHC.

A two-tailed Mann-Whitney $U$ test shows that the results are significant when comparing the worst average games of each algorithm ($p \ll 0.0001$ for N-Tuple over RMHC and B-RMHC). The differences between RMHC and B-RMHC are not statistically significant ($p = 0.5774$). If taking into account all $50$ trials, then N-Tuple remains significantly better than the other two algorithms ($p \ll 0.0001$). However, B-RMHC is not significantly better than RMHC ($p = 0.6080$).

Table \ref{bestgames} provides the parameters of the games, optimized by the RMHC, the B-RMHC and the N-Tuple algorithm, with the highest and lowest average fitness. 

\begin{table}
\centering
\caption{\label{bestgames} Optimized parameters of game instances with the highest or lowest average fitness, designed by three algorithms. 
}
\scriptsize
\resizebox{\columnwidth}{!}{%
\begin{tabular}{|c|cc|cc|cc|}
\hline
\multirow{3}{*}{Parameter} & \multicolumn{6}{c|}{Value optimised by different algorithms}\\
\cline{2-7}
&  \multicolumn{2}{c|}{RMHC} & \multicolumn{2}{c|}{B-RMHC} & \multicolumn{2}{c|}{N-Tuple}\\
\cline{2-7}
& High & \multicolumn{1}{c|}{Low} &High&\multicolumn{1}{c|}{Low}&High&Low\\
\hline
MISSILE\_MAX\_SPEED&6&1&10&1&9&10 \\
MISSILE\_COOLDOWN&9&5&5&3&2&5 \\
MISSILE\_RADIUS&2&10&10&4&4&4 \\
MISSILE\_MAX\_TTL&140&60&40&80&40&140 \\
GRID\_SIZE&4&3&1&1&3&1 \\
BLACKHOLE\_CELL(1,1)&0&1&0&1&1&1 \\
BLACKHOLE\_CELL(1,2)&0&0&1&1&0&1 \\
BLACKHOLE\_CELL(1,3)&0&1&1&0&0&1 \\
BLACKHOLE\_CELL(1,4)&1&0&0&0&0&0 \\
BLACKHOLE\_CELL(2,1)&0&0&1&1&0&1 \\
BLACKHOLE\_CELL(2,2)&1&1&0&1&0&0 \\
BLACKHOLE\_CELL(2,3)&1&0&1&0&0&1 \\
BLACKHOLE\_CELL(2,4)&1&1&1&0&1&0 \\
BLACKHOLE\_CELL(3,1)&1&1&1&0&0&0 \\
BLACKHOLE\_CELL(3,2)&1&0&1&0&1&1 \\
BLACKHOLE\_CELL(3,3)&1&0&0&0&1&1 \\
BLACKHOLE\_CELL(3,4)&1&1&0&0&1&1 \\
BLACKHOLE\_CELL(4,1)&1&0&0&1&0&0 \\
BLACKHOLE\_CELL(4,1)&1&0&0&0&0&1 \\
BLACKHOLE\_CELL(4,3)&1&0&0&1&0&1 \\
BLACKHOLE\_CELL(4,4)&0&1&0&0&1&1 \\
BLACKHOLE\_RADIUS&200&75&100&100&150&25 \\
BLACKHOLE\_FORCE&2&1&3&3&3&1 \\
BLACKHOLE\_PENALTY&3&4&0&7&7&8 \\
SAFE\_ZONE&20&0&20&20&10&10 \\
BOMB\_RADIUS&10&50&20&40&20&20 \\
MISSILE\_TYPE&2&1&2&0&2&0 \\
RESOURCE\_TTL&400&500&500&500&400&500 \\
RESOURCE\_COOLDOWN&200&250&250&200&200&200 \\
ENEMY\_ID&0&2&1&0&0&5 \\
\hline
\end{tabular}%
}
\end{table}

\subsection{Evaluation by human players}

\begin{figure*}[ptb]
\centering
\begin{subfigure}[t]{.32\textwidth}
\includegraphics[width=1\textwidth]{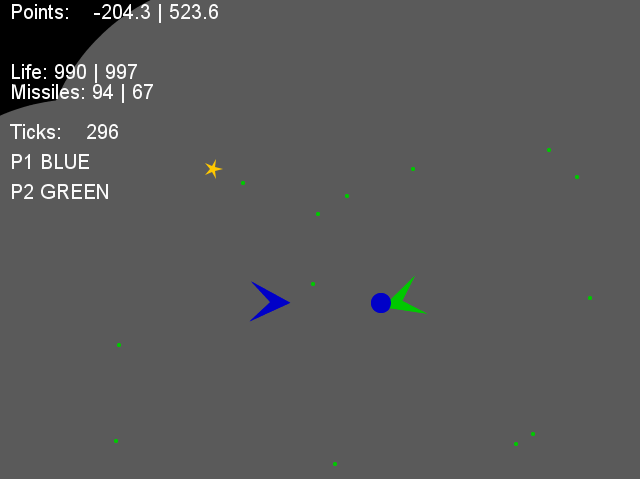}
\caption{$G1_H$, game with the highest average fitness designed by the RMHC
\bff{Player A} complained that although the environment was interesting, there are too many huge black holes overlapping, making it impossible to move. 
\bff{Player B} found this game annoying, as she was dragged into the center of black holes and couldn't escape, nor get resources.}
\label{fig:rmhcH} 
\end{subfigure}
\hfill
\begin{subfigure}[t]{0.32\textwidth}
\includegraphics[width=1\textwidth]{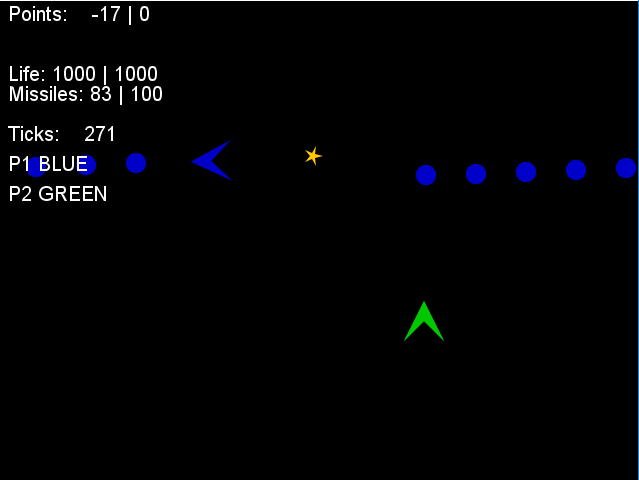}
\caption{$G2_H$, game with the highest average fitness designed by the Biased Mutation RMHC. \bff{Player A} found this game terrible (playing against doNothing controller with no black hole and huge missile bombs). \bff{Player B} found this game boring.}
\label{fig:mutH} 
\end{subfigure}
\hfill
\begin{subfigure}[t]{0.32\textwidth}
\includegraphics[width=1\textwidth]{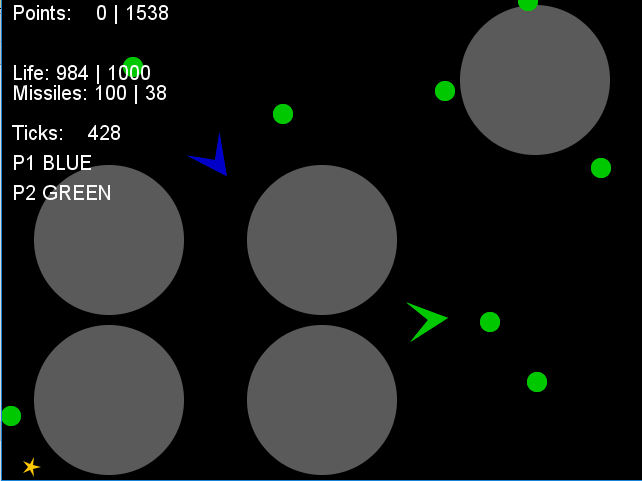}
\caption{$G3_H$, game with the highest average fitness designed by the N-Tuple Bandit Mutation. \bff{Player A} found this game interesting (with 2 big very strong black holes), but it was trivial because the ships could shoot each other when they were in the different black hole centers. \bff{Player B} found this game frustrating, but slightly better than $G3_L$.}
\label{fig:ntupleH} 
\end{subfigure}\\
\begin{subfigure}[t]{.32\textwidth}
\includegraphics[width=1\textwidth]{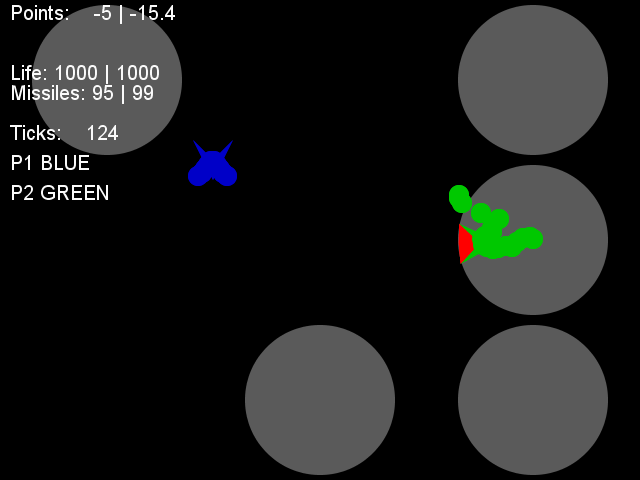}
\caption{$G1_L$, game with the lowest average fitness designed by the RMHC. \bff{Player A} found it difficult hitting the opponent, but the environment was dynamic and fun. \bff{Player B} found it better than $G1_H$, but still annoying, and suggested to increase the speed of missiles.}
\label{fig:rmhcL} 
\end{subfigure}
\hfill
\begin{subfigure}[t]{.32\textwidth}
\includegraphics[width=1\textwidth]{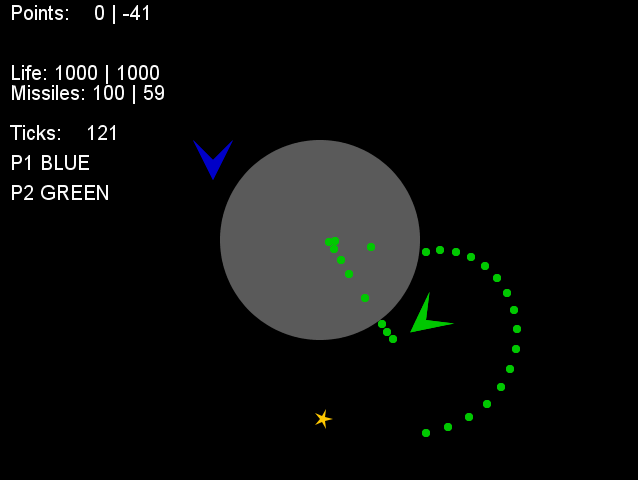}
\caption{$G2_L$, game with the lowest average fitness designed by the Biased Mutation RMHC. \bff{Player A} found this game not challenging at all. \bff{Player B} claimed that this game is her favorite.}
\label{fig:mutL} 
\end{subfigure}
\hfill
\begin{subfigure}[t]{.32\textwidth}
\includegraphics[width=1\textwidth]{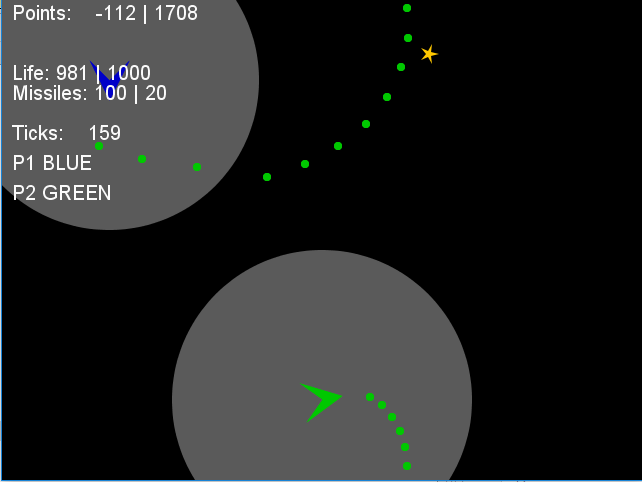}
\caption{$G3_L$, game with the lowest average fitness designed by the N-Tuple Bandit Mutation. \bff{Player A} found this game quite silly because of the missiles flying everywhere. \bff{Player B} found it good, though not as good as $G2_L$.}
\label{fig:ntupleL} 
\end{subfigure}
\caption{\label{screenshot}Screenshots of the 6 designed games evaluated by human players and their feedback. The players were not told which games they were playing. The game IDs were issued when analyzing the feedback. The \bff{Player A} ranked the games as $G1_L > G3_H > G1_H > G3_L> G2_H>G2_L$ in terms of challenge/fun, while the \bff{Player B} ranked the same games as $G2_L>G3_H>G3_L>G1_H>G1_L>G2_H$.}
\end{figure*}

We picked up the games with the highest and lowest average fitness designed by the 3 algorithms and invited two human players to evaluate them. The human players were asked to play the $6$ games and provide feedback without being told the fitness level of each game. One screenshot of each of the games is presented in Figure \ref{screenshot}, as well as the feedback from both players. The two human players evaluated the games differently, according to their playing preference. Player A cares more about the challenging aspect of the game and is attracted more towards uncommon game scenarios; Player B is less easily satisfied and found most of the games boring. Interestingly, though they have ranked the games differently, they both have a preference for the game $G3_H$ (with the highest average fitness value, optimised by N-Tuple) and dislike the games $G1_H$ (with the highest average fitness value, optimised by RMHC) and $G2_H$ (with the highest average fitness value, optimised by B-RMHC).

\subsection{Manual tuning of the evolved game}

It's notable that the game $G2_H$ uses a \emph{doNothing} opponent. We manually edited the ENEMY\_ID parameter to use an MCTS controller as opponent instead and asked the two human players to play the edited game. 
Player A found the new game improved but still a basic game with big missiles, not very interesting compared to the previous games.
However, Player B found the new game better with the agent now moving around the map and even increased its position in their personal ranking.


\section{Conclusion}\label{conc}

One of the big challenges in Game Design is the tuning of game parameters. Given a set of parameter values, a new game instance is created. The difficulty of a game could change significantly when varying one single parameter of a game. The behavior of a human player or an AI agent and the fun level of the game will also be affected. For instance, doubling the gravity in \emph{Flappy Bird} will expect to increase the frequency of calling the ``jump'' actions. However, the selection of game parameters and tuning  are not trivial due to the number of parameter to be tuned and the number of possible values of each of the parameters, resulting in a large search space. This motivates the research presented in this paper.

The authors applied the Random Mutation Hill Climber (RMHC) and two new algorithms, the Biased Mutation RMHC (B-RMHC) and the N-Tuple Bandit Evolutionary Algorithm (N-Tuple), to evolving game instances based on a real-time continuous 2-player competitive game called \emph{Space Battle Evolved} (detailed in \S\ref{newgame}). 

The Biased Mutation RMHC exploits some particular parameters which are considered to be more important after some pre-selection process. The N-Tuple Bandit Evolutionary Algorithm uses a bandit approach to balance the exploration and exploitation of the search space of every game parameter and a model to estimate the quality of unsampled game instances. The statistical results based on the final fitness of the solutions found by the three algorithms suggest the N-Tuple to be significantly better than the other two methods, being able to produce high fitness games.

Two human players have tested some of the evolved games and provided valuable reviews.
Both players preferred the new game evolved using the N-Tuple Bandit Evolutionary Algorithm, although they offered mixed opinions on the RMHC games. One highlight of this study is evolving the enemy AI as part of the game parameters. The effect of changing the opponent player was explored in the human trials, indicating that even though this aspect has a great effect on the quality of the gameplay, an outstandingly easy or difficult environment reduces this effect slightly. 

The experimental results on optimising \emph{Space Battle Evolved} (\S\ref{xp}) illustrate the outstanding and robust performance of the N-Tuple Bandit Evolutionary Algorithm. With this in mind, we can foresee a bright future for the N-Tuple Bandit Evolutionary Algorithm in AI-Assisted Game Design. Further work will look into the benefits of increasing the number of fitness evaluations, meant to reduce the noise in the evolution. Additionally, although the novel approach used in the Biased Mutation RMHC shows promise, improvements should be considered, such as increasing the re-sampling when measuring parameter importance metrics to produce more accurate results. Another possible future work is to apply this evolutionary algorithm with other game framework, such as GVG-AI \cite{GVGAI}.



\balance

\bibliographystyle{IEEEtran}
\bibliography{refs}

\begin{thebibliography}{10}
\providecommand{\url}[1]{#1}
\csname url@samestyle\endcsname
\providecommand{\newblock}{\relax}
\providecommand{\bibinfo}[2]{#2}
\providecommand{\BIBentrySTDinterwordspacing}{\spaceskip=0pt\relax}
\providecommand{\BIBentryALTinterwordstretchfactor}{4}
\providecommand{\BIBentryALTinterwordspacing}{\spaceskip=\fontdimen2\font plus
\BIBentryALTinterwordstretchfactor\fontdimen3\font minus
  \fontdimen4\font\relax}
\providecommand{\BIBforeignlanguage}[2]{{%
\expandafter\ifx\csname l@#1\endcsname\relax
\typeout{** WARNING: IEEEtran.bst: No hyphenation pattern has been}%
\typeout{** loaded for the language `#1'. Using the pattern for}%
\typeout{** the default language instead.}%
\else
\language=\csname l@#1\endcsname
\fi
#2}}
\providecommand{\BIBdecl}{\relax}
\BIBdecl

\bibitem{5035629}
J.~Togelius and J.~Schmidhuber, ``{An Experiment in Automatic Game Design},''
  in \emph{2008 IEEE Symposium On Computational Intelligence and Games}, Dec
  2008, pp. 111--118.

\bibitem{BanditRMHC}
\BIBentryALTinterwordspacing
J.~Liu, D.~P. Liebana, and S.~M. Lucas, ``{Bandit-Based Random Mutation
  Hill-Climbing},'' \emph{ArXiv}, 2016. [Online]. Available:
  \url{http://arxiv.org/abs/1606.06041}
\BIBentrySTDinterwordspacing

\bibitem{Automated:AIIA07}
M.~J. Nelson and M.~Mateas, ``{Towards Automated Game Design},'' in \emph{AI*IA
  2007: Artificial Intelligence and Human-Oriented Computing}.\hskip 1em plus
  0.5em minus 0.4em\relax Springer, 2007, pp. 626--637, lecture Notes in
  Computer Science 4733.

\bibitem{IsaksenGN15}
A.~Isaksen, D.~Gopstein, and A.~Nealen, ``{Exploring Game Space Using Survival
  Analysis},'' in \emph{Proceedings of the 10th International Conference on the
  Foundations of Digital Games, {FDG} 2015, Pacific Grove, CA, USA, June 22-25,
  2015}, 2015.

\bibitem{IsaksenICCC15}
A.~Isaksen, D.~Gopstein, J.~Togelius, and A.~Nealen, ``{Discovering Unique Game
  Variants},'' in \emph{Computational Creativity and Games Workshop, hosted by
  The Sixth International Conference on Computational Creativity, ICCC}, 2015.

\bibitem{AIIDE1511594}
A.~Isaksen and A.~Nealen, ``{Comparing Player Skill, Game Variants, and
  Learning Rates Using Survival Analysis},'' in \emph{AAAI Conference on
  Artificial Intelligence and Interactive Digital Entertainment}, 2015.

\bibitem{6605565}
D.~Perez, J.~Togelius, S.~Samothrakis, P.~Rohlfshagen, and S.~M. Lucas,
  ``{Automated Map Generation for the Physical Traveling Salesman Problem},''
  \emph{IEEE Transactions on Evolutionary Computation}, vol.~18, no.~5, pp.
  708--720, Oct 2014.

\bibitem{SafakPM16}
A.~B. Safak, E.~Bostanci, and A.~E. Soylucicek, ``{Automated Maze Generation
  for Ms. Pac-Man Using Genetic Algorithms},'' in \emph{{International Journal
  of Machine Learning and Computing}}, vol.~6, no.~4, 2016, pp. 226--240.

\bibitem{Togelius2011PCG}
K.~O.~S. Julian~Togelius, Georgios N.~Yannakakis and C.~Browne, ``{Search-based
  Procedural Content Generation: A Taxonomy and Survey},'' in \emph{IEEE
  Transactions on Computational Intelligence and AI in Games (TCIAIG)}, vol.~3,
  no.~3, 2011, pp. 172--186.

\bibitem{RothalaufEA}
F.~Rothlauf, \emph{Representations for Genetic and Evolutionary
  Algorithms}.\hskip 1em plus 0.5em minus 0.4em\relax Heidelberg: Springer,
  2006.

\bibitem{7317941}
T.~S. Nielsen, G.~A.~B. Barros, J.~Togelius, and M.~J. Nelson, ``{Towards
  Generating Arcade Game Rules with VGDL},'' in \emph{2015 IEEE Conference on
  Computational Intelligence and Games (CIG)}, Aug 2015, pp. 185--192.

\bibitem{4100132}
T.~L.~T. Menezes, T.~R. Baptista, and E.~J.~F. Costa, ``{Towards Generation of
  Complex Game Worlds},'' in \emph{2006 IEEE Symposium on Computational
  Intelligence and Games}, May 2006, pp. 224--229.

\bibitem{GVGAI}
D.~Perez-Liebana, S.~Samothrakis, J.~Togelius, T.~Schaul, S.~Lucas,
  A.~Couetoux, J.~Lee, C.-U. Lim, and T.~Thompson, ``{{The 2014 General Video
  Game Playing Competition}},'' in \emph{IEEE Transactions on Computational
  Intelligence and AI in Games}, vol.~PP, no.~99, 2015, p.~1.

\bibitem{GVGAI2P}
R.~D. Gaina, D.~Perez-Liebana, and S.~M. Lucas, ``{{General Video Game for 2
  Players: Framework and Competition}},'' in \emph{{Proceedings of the IEEE
  Computer Science and Electronic Engineering Conf.}}, 2016.

\bibitem{Liu2016}
J.~Liu, D.~Perez-Liebana, and S.~M. Lucas, ``{{Rolling Horizon Coevolutionary
  Planning for Two-Player Video Games}},'' in \emph{{Proceedings of the IEEE
  Conference on Computational intelligence and Games (CIG)}}, 2016, p. to
  appear.

\bibitem{MCTSsurvey}
C.~Browne, E.~Powley, D.~Whitehouse, S.~Lucas, P.~Cowling, P.~Rohlfshagen,
  S.~Tavener, D.~Perez, S.~Samothrakis, and S.~Colton, ``{{A Survey of Monte
  Carlo Tree Search Methods}},'' in \emph{{IEEE Trans. on Computational
  Intelligence and AI in Games}}, vol.~4, no.~1, 2014, pp. 1--43.

\bibitem{6784612}
M.~Buzdalov, A.~Buzdalova, and A.~Shalyto, ``{A First Step towards the Runtime
  Analysis of Evolutionary Algorithm Adjusted with Reinforcement Learning},''
  in \emph{2013 12th International Conference on Machine Learning and
  Applications}, vol.~1, Dec 2013, pp. 203--208.

\bibitem{NTupleOthello}
S.~M. Lucas, ``{Learning to Play Othello with N-Tuple Systems},''
  \emph{Australian Journal of Intelligent Information Processing}, vol.~4, pp.
  1--20, 2008.

\end{thebibliography}

\end{document}